\documentclass[runningheads]{llncs}
\usepackage{graphicx}    
\usepackage{epstopdf}
\usepackage{color}
\usepackage{cite}
\usepackage{amsfonts,amssymb,amsmath}
\usepackage{multirow}
\usepackage{epsf}
\usepackage{longtable}
\usepackage{booktabs}
\usepackage[table]{xcolor}
\usepackage{ragged2e}
\usepackage{subfigure}
\usepackage{marvosym}
\usepackage{rotating}
\usepackage{url}

\begin{document}
%
\title{SAM-U: Multi-box prompts triggered uncertainty estimation for reliable SAM in medical image}
\author{Guoyao Deng\inst{1}, Ke Zou\inst{1,3}, Kai Ren\inst{2}, Meng Wang\inst{3}, Xuedong Yuan\inst{2}, Sancong Ying\inst{2} and Huazhu Fu\inst{3}} 
\authorrunning{G.Deng et al.}
\institute{National Key Laboratory of Fundamental Science on Synthetic Vision,\\ Sichuan
University, Sichuan, China \and College of Computer Science, Sichuan University, Sichuan, China \and  Institute of High Performance Computing, A*STAR, Singapore
}
\maketitle
\begin{abstract}
Recently, Segmenting Anything Model has taken a significant step towards general artificial intelligence. Simultaneously, its reliability and fairness have garnered significant attention, particularly in the field of healthcare. In this study, we propose a multi-box prompt-triggered uncertainty estimation for SAM cues to demonstrate the reliability of segmented lesions or tissues. We estimate the distribution of SAM predictions using Monte Carlo with prior distribution parameters, employing different prompts as a formulation of test-time augmentation. Our experimental results demonstrate that multi-box prompts augmentation enhances SAM performance and provides uncertainty for each pixel. This presents a groundbreaking paradigm for a reliable SAM.


\end{abstract}
\section{Introduction}
\label{sec:introduction}
Large-scale foundation models are increasingly gaining popularity among artificial intelligence researchers. In the realm of natural language processing (NLP), the Generative Pre-trained Transformer (GPT)~\cite{floridi2020gpt} and ChatGPT, developed by OpenAI, have witnessed rapid growth owing to their exceptional ability to generalize. These models have found applications in diverse domains such as autonomous driving and healthcare. The remarkable generalization capabilities of large models often instill a sense of trust among users; however, their fairness and reliability have also been subject to some degree of scrutiny.

Nowadays, there is a growing wave of enthusiasm surrounding computer vision due to the release of the Segment Anything Model (SAM) ~\cite{kirillov2023segment}by Meta AI. SAM has been trained on a massive SA-1B dataset, which consists of over 11 million images and one billion masks, making it an excellent tool. It excels at producing accurate segmentation results from various types of prompts, including foreground/background points, thick boxes or masks, and free-form text. The introduction of SAM has led many researchers to believe that general artificial intelligence has finally arrived. However, some researchers have expressed concerns about the performance of SAM~\cite{tang2023can,ji2023sam,ji2023segment,he2023can,roy2023sam}. Specifically, they have identified areas such as industrial defect detection~\cite{ji2023sam,ji2023segment}, camouflaged target detection~\cite{ji2023sam,ji2023segment,tang2023can}, and tumor and lesion segmentation ~\cite{tang2023can,he2023can,roy2023sam}in medical images where further improvements are needed. Additionally, the reliability of SAM still requires further study.

Uncertainty estimation~\cite{16dropout} is one of the ways to provide reliability for SAM. Previously, uncertainty estimation has demonstrated its reliability and robustness in several medical segmentation tasks~\cite{zou2022tbrats,zou2023evidencecap}, including skin lesions and brain tumors~\cite{li2022region}, among others. The current uncertainty estimation methods can be roughly divided into deterministic-based methods~\cite{ICMLdeterministic20,2020NIPSdeterministic}, Bayesian Neural Network-based methods~\cite{roy2019bayesian}, ensemble-based methods~\cite{ensemble17}, dropout-based methods~\cite{16dropout,17dropoutCV,MIA2020exploringDropSeg} and test-time augmentation-based methods~\cite{wang2018TTA}. The focus of this paper is to keep the simplicity and retain the original structure of SAM while achieving pixel-level uncertainty estimation.

\begin{figure}[!htbp]
\centering
\includegraphics[width=1\linewidth]{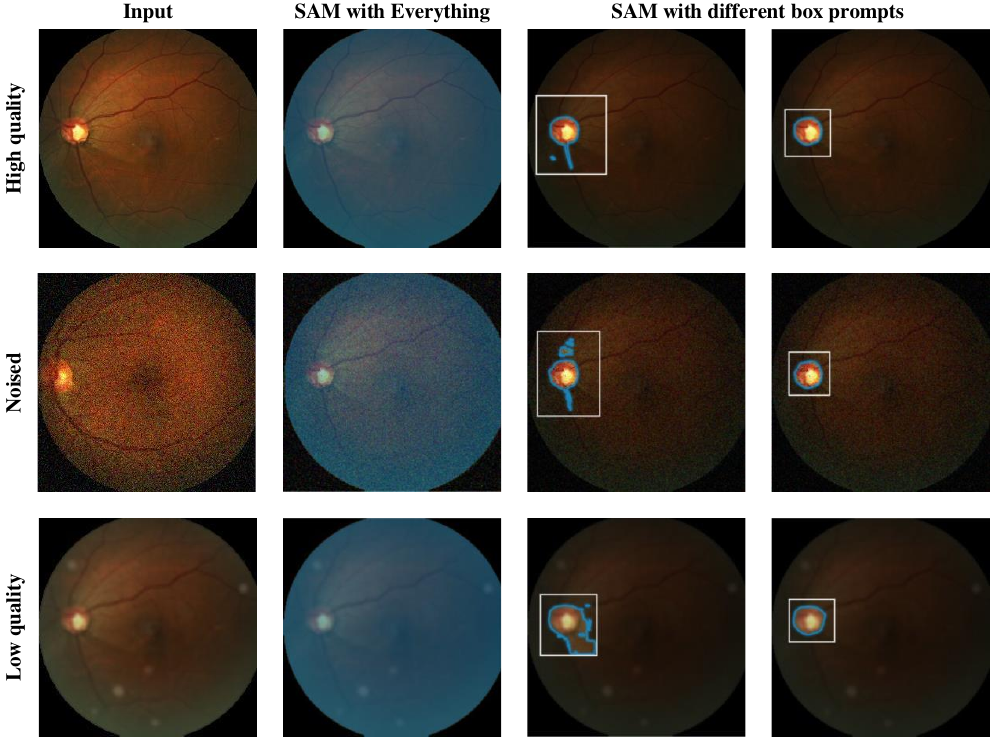}
\caption{Prediction results of different quality fundus images in SAM's everything and box modes.}
\label{F_1}
\end{figure}

In Fig.\ref{F_1}, we present the eye disc segmentation results\cite{fu2018joint} for both high and low-quality fundus images under different conditions. SAM demonstrates better segmentation results for high-quality images, and the inclusion of different conditions leads to certain performance improvements. However, SAM's segmentation results for lower quality images are not satisfactory. Nevertheless, the inclusion of different conditions greatly enhances its performance, particularly with more accurate box prompts. Furthermore, we have observed a phenomenon wherein different levels of box prompts tend to yield diverse results. This observation motivates us to introduce a novel approach, namely multi-box prompts-induced uncertainty estimation, for medical images.

Therefore, the primary focus of this paper is to enhance the segmentation accuracy by employing multiple box prompts. This approach enables us to establish pixel-level reliability through uncertainty estimation. Specifically, we utilize SAM to predict the output distribution using different multi-box prompts. SAM with multi-box prompts generates numerous samples from the predictive distribution. Subsequently, these samples are used to calculate variance, which provide an uncertainty estimation for the medical image segmentation. Our experiments demonstrate that multi-box prompts not only enhances performance on low-quality medical images but also provides uncertainty estimation for them. 

\section{Method}
\begin{figure}[!htbp]
\centering
\includegraphics[width=1\linewidth]{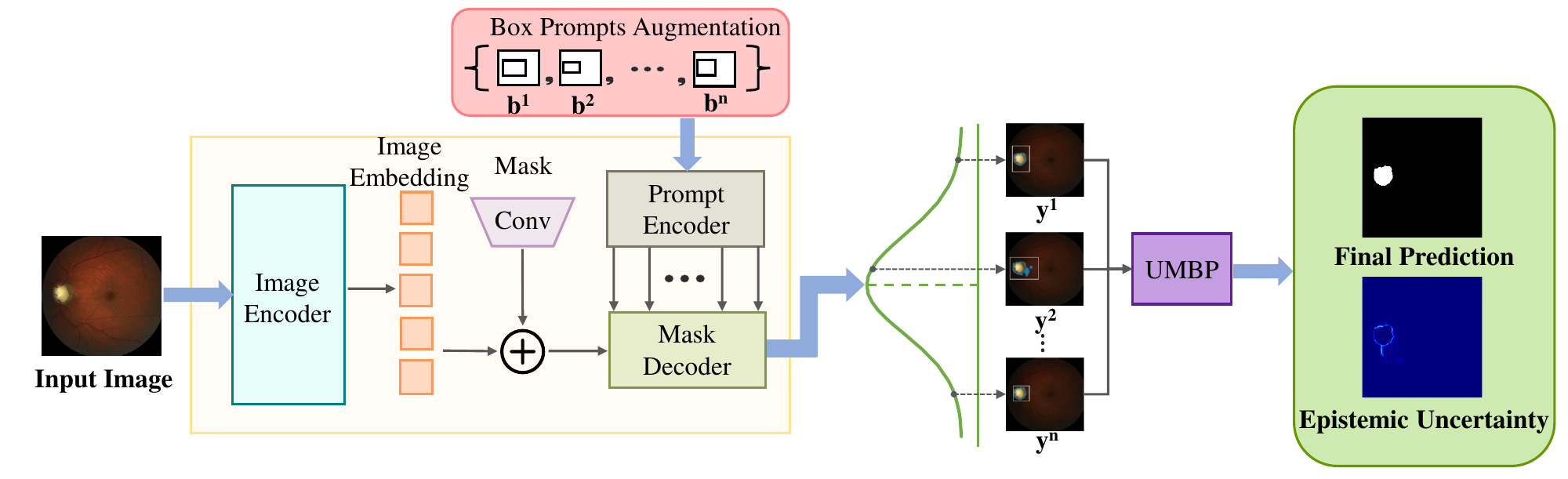}
\caption{The overall framework of SAM-U.}
\label{F_2}
\end{figure}

The overall framework of our proposed method is depicted in Fig.~\ref{F_2}. Our main focus is to enhance the reliability and accuracy of SAM in the context of zero-shot learning. To improve the accuracy of SAM, we incorporate multi-box prompts, which enable us to obtain more precise medical image segmentation results from the distribution. Specifically, we estimate the distribution of SAM predictions using Monte Carlo simulation with prior distribution parameters. This approach allows our method to estimate the aleatoric uncertainty by considering multiple forecasts for a single medical image.

\subsection{Mask Selection Strategy}
Under the unprompted setting, SAM generates multiple binary masks and can pop out several potential objects within an input. For a fair evaluation of interesting regions in a specific segmentation task, we follow the strategy of \cite{ji2023sam}to select the most appropriate mask based on its ground-truth mask. Formally, given $N$ binary predictions $\{y^i\}_i^N=1$ and the ground-truth $G$ for an input image, we calculate Dice scores for each pair to generate a set of evaluation scores $\{D^i\}_i^N=1$. We finally select the mask with the highest $Dice$ score from this set.

\subsection{SAM with multi-box prompts}
Prompts can introduce errors into the model's inferring due to their inherent inaccuracies. In order to reduce the influence of the variance of the prompt. We randomize $M$ box prompts $B=\{b^1,b^2,\cdots,b^M\}$. Each  box prompt guides SAM generates different segmentation results. Through this strategy, we obtain the predictions $Y=\{y^1,y^2,\cdots,y^M\}$ of SAM under different prior cues, and combining them can improve the segmentation accuracy of SAM and reduce uncertainty. The combined prediction is computed as:
\begin{equation}
\hat y = \frac{1}{M}\sum\limits_{i = 1}^M {{f_{SAM}}\left( {I,{b^i}} \right)} ,  
\label{E_1}
\end{equation}
where $ y_C$ denotes the combined prediction of image I.
\subsection{Uncertainty estimation of SAM with multi-box prompts}
\begin{figure}[!t]
\centering
\includegraphics[width=0.6\linewidth]{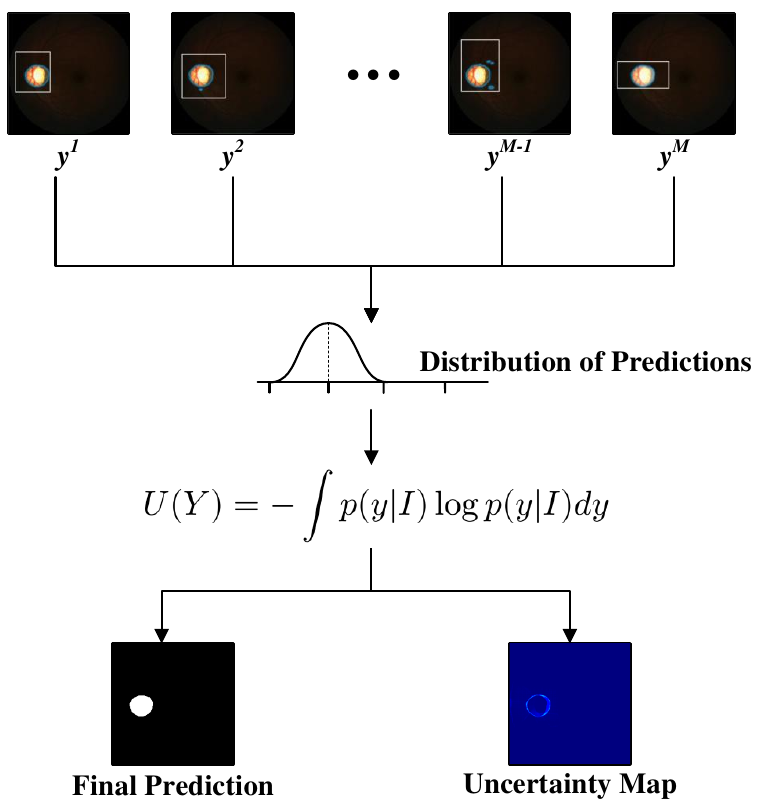}
\caption{The uncertainty distribution approximated by Monte Carlo Simulation.}
\label{F_3}
\end{figure}

Different box prompts cause variances in SAM's segmentation even if they refer to one object in human's view. Inspired by this, our proposed multi-box prompts (MNP) algorithm simulates the annotations of multiple clinical experts to generate the final predictions and uncertainty estimations. To quantify the uncertainty triggered by multi-box prompts. Assume $M$ box prompts $B=\{b^1,b^2,\cdots,b^M\}$ that all refers to the ground truth. With $M$ box prompts and input image $I$, SAM generate a set of predictions $Y=\{y^1,y^2,\cdots,y^M\}$.  As shown in Fig.~\ref{F_3}, We present an uncertainty estimation procedure for multi-box prompts.

We first describe aleatoric uncertainty from a single given image $I$ by the entropy~\cite{bein2006entropy}:
\begin{equation}
U(y^i) = -\int p(y^i|I) \log{p(y^i|I)} dy,  
\label{E_2}
\end{equation}
$U(y^i)$ estimates how diverse the prediction ${y^i}$ for the image $I$. where ${y^i} = \{ p_{_1}^i,p_{_2}^i, \cdots ,p_{_N}^i\} $ denote the prediction pixels. $N$ denotes the unique values in ${y^i}$.  

Then,  We run a Monte Carlo simulation using multi-box prompts to obtain a set of predictions. Therefore, the uncertainty distribution is approximated as follows:
\begin{equation}
U(Y|I) \approx \sum\limits_{i = 1}^M {\sum\limits_{j = 1}^N {p_j^i\log } } p_j^i, 
\label{E_3}
\end{equation}

\section{Experiments and Results}
Two different methods are utilized to perform image degradation to verify the reliability of SAM. In this section, we will describe our evaluation protocols, compare the performance of SAM under different quality datasets, and visualize the qualitative results on fundus image segmentation.

\subsection{Evaluation Protocols}
$\bullet$  \textbf{Dataset.} We chose the sub-task of the REFUGE Challenge ~\cite{orlando2020refuge}, which does the segmentation of  the optic cup and disc in fundus photographs. For simplicity's sake, we consider disc and cup as one category. In order to evaluate the reliability of SAM more objectively, we artificially constructed low-quality data based on high-quality source data by two different methods, which is introducing Gaussian noise with various levels of standard deviations ($\sigma$) and the realistic degradation model proposed by \textit{Shen et al.}~\cite{deep_reitna_enhance}, respectively.\\
$\bullet$  \textbf{Metrics.} We use four commonly-used metrics for the evaluation: dice score (Dice), expected calibration error (ECE)~\cite{guo2017calibration}, structure measure (Sm)~\cite{fan2017structure} and weighted F-measure (wFm)~\cite{margolin2014evaluate}.

\subsection{Quantitative Evaluation}

\begin{table}[!t]
  \centering
  \caption{Quantitative results of high-quality fundus images in SAM's everything and box modes. SAM$^1$, SAM$^2$ and SAM$^3$ mean using Everything, Box, and our modes, respectively.\label{T1}}
    \begin{tabular}{ccccccc}
    \toprule
    \rowcolor{gray} Model & Backbone & Dice$\uparrow$  & ECE$\downarrow$ & Sm$\uparrow$    & wFm$\uparrow$  \\
    \midrule
    \multirow{3}[1]{*}{SAM$^1$} & ViT-B &  0.436     &  0.050   &  0.394     &  0.436      \\
          & ViT-L &   0.438   &   0.050   &   0..394 &   0.438   \\
          & ViT-H &    0.441  &    0.049   &  0.394   &    0.441  \\
    \hline\hline
    \multirow{6}[1]{*}{SAM$^2$} & ViT-B &  0.533  &   0.016 &   0.404  & 0.533   \\
          & \cellcolor[rgb]{ .647,  .647,  .647}{Difference ($\Delta$)} &   \cellcolor[rgb]{ .647,  .647,  .647}{0.097}   &   \cellcolor[rgb]{ .647,  .647,  .647}\textcolor[rgb]{ 1,  0,  0}{(0.034)}  & \cellcolor[rgb]{ .647,  .647,  .647}{0.01}     &  \cellcolor[rgb]{ .647,  .647,  .647}{ 0.097  }    \\
          & ViT-L &    0.739    &   0.014&  0.408    &  0.739  \\
          & \cellcolor[rgb]{ .647,  .647,  .647}{Difference ($\Delta$) }&  \cellcolor[rgb]{ .647,  .647,  .647}{0.301}     &  \cellcolor[rgb]{ .647,  .647,  .647}\textcolor[rgb]{ 1,  0,  0}{ (0.036)}    &  \cellcolor[rgb]{ .647,  .647,  .647}{ 0.014}    &     \cellcolor[rgb]{ .647,  .647,  .647}{0.301}     \\
          & ViT-H   &  0.691 &   0.020 &   0.406  & 0.691   \\
          & \cellcolor[rgb]{ .647,  .647,  .647}{Difference ($\Delta$)} &   \cellcolor[rgb]{ .647,  .647,  .647}{0.25}    &  \cellcolor[rgb]{ .647,  .647,  .647}\textcolor[rgb]{ 1,  0,  0}{ (0.029) }  &  \cellcolor[rgb]{ .647,  .647,  .647}{ 0.012}   &  \cellcolor[rgb]{ .647,  .647,  .647}{0.25 }     \\
    \hline\hline
    \multirow{6}[1]{*}{SAM$^3$} & ViT-B & 0.538 &  0.016 &   0.404  & 0.538 \\
          & \cellcolor[rgb]{ .647,  .647,  .647}{Difference ($\Delta$)} &   \cellcolor[rgb]{ .647,  .647,  .647}{0.102}   &    \cellcolor[rgb]{ .647,  .647,  .647}\textcolor[rgb]{ 1,  0,  0}{(0.034)}  &   \cellcolor[rgb]{ .647,  .647,  .647}{0.010}     &  \cellcolor[rgb]{ .647,  .647,  .647}{ 0.102 }    \\
          & ViT-L &    0.751   &   0.013 & 0.408    &   0.751  \\
          & \cellcolor[rgb]{ .647,  .647,  .647}{Difference ($\Delta$) }&  \cellcolor[rgb]{ .647,  .647,  .647}{0.313}     &  \cellcolor[rgb]{ .647,  .647,  .647}\textcolor[rgb]{ 1,  0,  0}{ (0.037)}    &  \cellcolor[rgb]{ .647,  .647,  .647}{ 0.014 }    &      \cellcolor[rgb]{ .647,  .647,  .647}{0.313}     \\
          & ViT-H &  0.718 &   0.017 &   0.407  & 0.718    \\
          & \cellcolor[rgb]{ .647,  .647,  .647}{Difference ($\Delta$)} &   \cellcolor[rgb]{ .647,  .647,  .647}{0.277}    &  \cellcolor[rgb]{ .647,  .647,  .647}\textcolor[rgb]{ 1,  0,  0}{ (0.032) }  &  \cellcolor[rgb]{ .647,  .647,  .647}{ 0.013}   &   \cellcolor[rgb]{ .647,  .647,  .647}{0.277 }     \\
    \bottomrule
    \end{tabular}%
  \label{tab:addlabel}%
\end{table}%
\begin{table}[!t]
  \centering
  \caption{Results on two different situations in which we add Gaussian Noise that $\sigma =0.05$ and $ \sigma=0.10$.\label{T2}}
    \begin{tabular}{cccccccccccccccccc}
    \toprule
    \multicolumn{2}{c}{\multirow{2}[4]{*}{Model}} & \multirow{2}[4]{*}{BackBone} & \multicolumn{7}{c}{$\sigma=0.05$ }       &    & \multicolumn{7}{c}{$\sigma=0.10$ } \\
\cmidrule{4-10}\cmidrule{12-18}    \multicolumn{2}{c}{} &    & Dice &    & ECE &    & SM &    & FM &    & Dice &    & ECE &    & SM &    & FM \\
\cmidrule{1-4}\cmidrule{6-6}\cmidrule{8-8}\cmidrule{10-10}\cmidrule{12-12}\cmidrule{14-14}\cmidrule{16-16}\cmidrule{18-18}    \multicolumn{2}{c}{\multirow{3}[2]{*}{SAM$^1$}} & ViT\_B & 0.293  &    & 0.180  &    & 0.552  &    & 0.293  &    & 0.214  &    & 0.318  &    & 0.447  &    & 0.214  \\
    \multicolumn{2}{c}{} & ViT\_L & 0.461  &    & 0.106  &    & 0.667  &    & 0.461  &    & 0.387  &    & 0.266  &    & 0.562  &    & 0.387  \\
    \multicolumn{2}{c}{} & ViT\_H & 0.496  &    & 0.109  &    & 0.680  &    & 0.496  &    & 0.427  &    & 0.171  &    & 0.622  &    & 0.427  \\
    \midrule
    \midrule
    \multicolumn{2}{c}{\multirow{6}[2]{*}{SAM$^2$}} & ViT\_B & 0.397  &    & 0.025  &    & 0.621  &    & 0.397  &    & 0.375  &    & 0.030  &    & 0.610  &    & 0.375  \\
    \multicolumn{2}{c}{} & \cellcolor[rgb]{ .647,  .647,  .647}Difference($\Delta$) & \cellcolor[rgb]{ .647,  .647,  .647}0.104  & \cellcolor[rgb]{ .647,  .647,  .647}  & \cellcolor[rgb]{ .647,  .647,  .647}\textcolor[rgb]{ 1,  0,  0}{(0.155)} & \cellcolor[rgb]{ .647,  .647,  .647}  & \cellcolor[rgb]{ .647,  .647,  .647}0.069  & \cellcolor[rgb]{ .647,  .647,  .647}  & \cellcolor[rgb]{ .647,  .647,  .647}0.104  & \cellcolor[rgb]{ .647,  .647,  .647} & \cellcolor[rgb]{ .647,  .647,  .647}0.161  & \cellcolor[rgb]{ .647,  .647,  .647}  & \cellcolor[rgb]{ .647,  .647,  .647}\textcolor[rgb]{ 1,  0,  0}{(0.288)} & \cellcolor[rgb]{ .647,  .647,  .647}& \cellcolor[rgb]{ .647,  .647,  .647}0.163  & \cellcolor[rgb]{ .647,  .647,  .647}  & \cellcolor[rgb]{ .647,  .647,  .647}0.161  \\
    \multicolumn{2}{c}{} & ViT\_L & 0.700  &    & 0.014  &    & 0.792  &    & 0.700  &    & 0.659  &    & 0.018  &    & 0.761  &    & 0.659  \\
    \multicolumn{2}{c}{} & \cellcolor[rgb]{ .647,  .647,  .647}Difference($\Delta$) & \cellcolor[rgb]{ .647,  .647,  .647}0.239  & \cellcolor[rgb]{ .647,  .647,  .647} & \cellcolor[rgb]{ .647,  .647,  .647}\textcolor[rgb]{ 1,  0,  0}{(0.092)} & \cellcolor[rgb]{ .647,  .647,  .647} & \cellcolor[rgb]{ .647,  .647,  .647}0.125  & \cellcolor[rgb]{ .647,  .647,  .647}  & \cellcolor[rgb]{ .647,  .647,  .647}0.239  & \cellcolor[rgb]{ .647,  .647,  .647} & \cellcolor[rgb]{ .647,  .647,  .647}0.272  & \cellcolor[rgb]{ .647,  .647,  .647}  & \cellcolor[rgb]{ .647,  .647,  .647}\textcolor[rgb]{ 1,  0,  0}{(0.248)} & \cellcolor[rgb]{ .647,  .647,  .647} & \cellcolor[rgb]{ .647,  .647,  .647}0.199  & \cellcolor[rgb]{ .647,  .647,  .647}  & \cellcolor[rgb]{ .647,  .647,  .647}0.272  \\
    \multicolumn{2}{c}{} & ViT\_H & 0.685  &    & 0.017  &    & 0.799  &    & 0.685  &    & 0.665  &    & 0.020  &    & 0.789  &    & 0.665  \\
    \multicolumn{2}{c}{} & \cellcolor[rgb]{ .647,  .647,  .647}Difference($\Delta$) & \cellcolor[rgb]{ .647,  .647,  .647}0.189  & \cellcolor[rgb]{ .647,  .647,  .647} & \cellcolor[rgb]{ .647,  .647,  .647}\textcolor[rgb]{ 1,  0,  0}{(0.092)} & \cellcolor[rgb]{ .647,  .647,  .647}  & \cellcolor[rgb]{ .647,  .647,  .647}0.119  & \cellcolor[rgb]{ .647,  .647,  .647}  & \cellcolor[rgb]{ .647,  .647,  .647}0.189  & \cellcolor[rgb]{ .647,  .647,  .647} & \cellcolor[rgb]{ .647,  .647,  .647}0.278  & \cellcolor[rgb]{ .647,  .647,  .647}  & \cellcolor[rgb]{ .647,  .647,  .647}\textcolor[rgb]{ 1,  0,  0}{(0.246)} & \cellcolor[rgb]{ .647,  .647,  .647}  & \cellcolor[rgb]{ .647,  .647,  .647}0.227  & \cellcolor[rgb]{ .647,  .647,  .647}  & \cellcolor[rgb]{ .647,  .647,  .647}0.278  \\
    \midrule
    \midrule
    \multicolumn{2}{c}{\multirow{6}[2]{*}{SAM$^3$}} & ViT\_B & 0.416  &    & 0.022  &    & 0.633  &    & 0.417  &    & 0.399  &    & 0.025  &    & 0.621  &    & 0.399  \\
    \multicolumn{2}{c}{} & \cellcolor[rgb]{ .647,  .647,  .647}Difference($\Delta$) & \cellcolor[rgb]{ .647,  .647,  .647}0.123  & \cellcolor[rgb]{ .647,  .647,  .647}  & \cellcolor[rgb]{ .647,  .647,  .647}\textcolor[rgb]{ 1,  0,  0}{(0.158)} & \cellcolor[rgb]{ .647,  .647,  .647}  & \cellcolor[rgb]{ .647,  .647,  .647}0.081  & \cellcolor[rgb]{ .647,  .647,  .647}  & \cellcolor[rgb]{ .647,  .647,  .647}0.124  & \cellcolor[rgb]{ .647,  .647,  .647} & \cellcolor[rgb]{ .647,  .647,  .647}0.185  & \cellcolor[rgb]{ .647,  .647,  .647} & \cellcolor[rgb]{ .647,  .647,  .647}\textcolor[rgb]{ 1,  0,  0}{(0.293)} & \cellcolor[rgb]{ .647,  .647,  .647}  & \cellcolor[rgb]{ .647,  .647,  .647}0.174  & \cellcolor[rgb]{ .647,  .647,  .647} & \cellcolor[rgb]{ .647,  .647,  .647}0.185  \\
    \multicolumn{2}{c}{} & ViT\_L & 0.726  &    & 0.012  &    & 0.808  &    & 0.726  &    & 0.690  &    & 0.015  &    & 0.779  &    & 0.690  \\
    \multicolumn{2}{c}{} & \cellcolor[rgb]{ .647,  .647,  .647}Difference($\Delta$) & \cellcolor[rgb]{ .647,  .647,  .647}0.265  & \cellcolor[rgb]{ .647,  .647,  .647} & \cellcolor[rgb]{ .647,  .647,  .647}\textcolor[rgb]{ 1,  0,  0}{(0.094)} & \cellcolor[rgb]{ .647,  .647,  .647} & \cellcolor[rgb]{ .647,  .647,  .647}0.141  & \cellcolor[rgb]{ .647,  .647,  .647}  & \cellcolor[rgb]{ .647,  .647,  .647}0.265  & \cellcolor[rgb]{ .647,  .647,  .647} & \cellcolor[rgb]{ .647,  .647,  .647}0.303  & \cellcolor[rgb]{ .647,  .647,  .647}  & \cellcolor[rgb]{ .647,  .647,  .647}\textcolor[rgb]{ 1,  0,  0}{(0.251)} & \cellcolor[rgb]{ .647,  .647,  .647}  & \cellcolor[rgb]{ .647,  .647,  .647}0.217  & \cellcolor[rgb]{ .647,  .647,  .647}  & \cellcolor[rgb]{ .647,  .647,  .647}0.303  \\
    \multicolumn{2}{c}{} & ViT\_H & 0.711  &    & 0.014  &    & 0.814  &    & 0.711  &    & 0.710  &    & 0.016  &    & 0.817  &    & 0.710  \\
    \multicolumn{2}{c}{} & \cellcolor[rgb]{ .647,  .647,  .647}Difference($\Delta$) & \cellcolor[rgb]{ .647,  .647,  .647}0.215  & \cellcolor[rgb]{ .647,  .647,  .647}  & \cellcolor[rgb]{ .647,  .647,  .647}\textcolor[rgb]{ 1,  0,  0}{(0.095)} & \cellcolor[rgb]{ .647,  .647,  .647} & \cellcolor[rgb]{ .647,  .647,  .647}0.134  & \cellcolor[rgb]{ .647,  .647,  .647}  & \cellcolor[rgb]{ .647,  .647,  .647}0.215  & \cellcolor[rgb]{ .647,  .647,  .647} & \cellcolor[rgb]{ .647,  .647,  .647}0.283  & \cellcolor[rgb]{ .647,  .647,  .647}  & \cellcolor[rgb]{ .647,  .647,  .647}\textcolor[rgb]{ 1,  0,  0}{(0.155)} & \cellcolor[rgb]{ .647,  .647,  .647}  & \cellcolor[rgb]{ .647,  .647,  .647}0.195  & \cellcolor[rgb]{ .647,  .647,  .647} & \cellcolor[rgb]{ .647,  .647,  .647}0.283  \\
\cmidrule{1-2}    \end{tabular}%
  \label{tab:addlabel}%
\end{table}%

As shown in Table ~\ref{T1}, we present different segmentation results of SAM modes using high-quality medical images. Initially, we compare the segmentation results of SAM in "everything" mode and SAM in "box" mode on normal medical images. It was found that the results using SAM in "box" mode were superior. Moreover, with the introduction of our algorithm, the performance of SAM improved further. Table ~\ref{T2} and Table ~\ref{T3} demonstrate various segmentation results of SAM modes under Gaussian noise and degraded medical images. We compare the results obtained from the aforementioned SAM modes. The performance of SAM in "everything" mode and SAM in "box" mode has declined, whereas the performance of SAM with "multi-box" mode remains at a certain level, with a lower ECE index. Therefore, it can be concluded that the inclusion of multi-box prompts enhances the accuracy and reliability of SAM.

\begin{table}[!t]
  \centering
  \caption{Results on LQ dataset. The realistic degradation model proposed by\textit{ Shen et.}~\cite{deep_reitna_enhance} can generate 7 kinds of low-quality images from '000' to '111' based on the high-quality image. We chose the two that look most different from the original image to the naked eye.\label{T3}}
    \begin{tabular}{cccccccccccccccccc}
    \toprule
    \multicolumn{2}{c}{\multirow{2}[4]{*}{Model}} & \multirow{2}[4]{*}{BackBone} & \multicolumn{7}{c}{"101"}          &    & \multicolumn{7}{c}{"111"} \\
\cmidrule{4-10}\cmidrule{12-18}    \multicolumn{2}{c}{} &    & Dice &    & ECE &    & SM &    & FM &    & Dice &    & ECE &    & SM &    & FM \\
\cmidrule{1-4}\cmidrule{6-6}\cmidrule{8-8}\cmidrule{10-10}\cmidrule{12-12}\cmidrule{14-14}\cmidrule{16-16}\cmidrule{18-18}    \multicolumn{2}{c}{\multirow{3}[2]{*}{SAM$^1$}} & ViT\_B & 0.386  &    & 0.222  &    & 0.578  &    & 0.386  &    & 0.396  &    & 0.078  &    & 0.646  &    & 0.396  \\
    \multicolumn{2}{c}{} & ViT\_L & 0.470  &    & 0.130  &    & 0.663  &    & 0.470  &    & 0.465  &    & 0.075  &    & 0.685  &    & 0.465  \\
    \multicolumn{2}{c}{} & ViT\_H & 0.520  &    & 0.126  &    & 0.690  &    & 0.520  &    & 0.505  &    & 0.087  &    & 0.698  &    & 0.505  \\
    \midrule
    \midrule
    \multicolumn{2}{c}{\multirow{6}[2]{*}{SAM$^2$}} & ViT\_B & 0.542  &    & 0.019  &    & 0.713  &    & 0.542  &    & 0.541  &    & 0.021  &    & 0.715  &    & 0.541  \\
    \multicolumn{2}{c}{} & \cellcolor[rgb]{ .647,  .647,  .647}Difference($\Delta$) & \cellcolor[rgb]{ .647,  .647,  .647}0.156  & \cellcolor[rgb]{ .647,  .647,  .647} & \cellcolor[rgb]{ .647,  .647,  .647}\textcolor[rgb]{ 1,  0,  0}{(0.203)} & \cellcolor[rgb]{ .647,  .647,  .647} & \cellcolor[rgb]{ .647,  .647,  .647}0.135  & \cellcolor[rgb]{ .647,  .647,  .647} & \cellcolor[rgb]{ .647,  .647,  .647}0.156  & \cellcolor[rgb]{ .647,  .647,  .647} & \cellcolor[rgb]{ .647,  .647,  .647}0.145  & \cellcolor[rgb]{ .647,  .647,  .647}  & \cellcolor[rgb]{ .647,  .647,  .647}\textcolor[rgb]{ 1,  0,  0}{(0.057)} & \cellcolor[rgb]{ .647,  .647,  .647} & \cellcolor[rgb]{ .647,  .647,  .647}0.069  & \cellcolor[rgb]{ .647,  .647,  .647}  & \cellcolor[rgb]{ .647,  .647,  .647}0.145  \\
    \multicolumn{2}{c}{} & ViT\_L & 0.679  &   & 0.018  &   & 0.773  &    & 0.679  &    & 0.684  &    & 0.020  &    & 0.778  &    & 0.684  \\
    \multicolumn{2}{c}{} & \cellcolor[rgb]{ .647,  .647,  .647}Difference($\Delta$) & \cellcolor[rgb]{ .647,  .647,  .647}{0.209} & \cellcolor[rgb]{ .647,  .647,  .647} & \cellcolor[rgb]{ .647,  .647,  .647}\textcolor[rgb]{ 1,  0,  0}{(0.012)} & \cellcolor[rgb]{ .647,  .647,  .647}  & \cellcolor[rgb]{ .647,  .647,  .647}{0.110} &  \cellcolor[rgb]{ .647,  .647,  .647} & \cellcolor[rgb]{ .647,  .647,  .647}{0.209} & \cellcolor[rgb]{ .647,  .647,  .647} & \cellcolor[rgb]{ .647,  .647,  .647}0.219  & \cellcolor[rgb]{ .647,  .647,  .647} & \cellcolor[rgb]{ .647,  .647,  .647}\textcolor[rgb]{ 1,  0,  0}{(0.055)} & \cellcolor[rgb]{ .647,  .647,  .647}  & \cellcolor[rgb]{ .647,  .647,  .647}0.093  & \cellcolor[rgb]{ .647,  .647,  .647} & \cellcolor[rgb]{ .647,  .647,  .647}0.219  \\
    \multicolumn{2}{c}{} & ViT\_H &0.596    &    & 0.031  &    & 0.746  &    & 0.596  &    & 0.589  &    & 0.032  &    & 0.742  &    & 0.589  \\
    \multicolumn{2}{c}{} & \cellcolor[rgb]{ .647,  .647,  .647}Difference($\Delta$) & \cellcolor[rgb]{ .647,  .647,  .647}{0.076} & \cellcolor[rgb]{ .647,  .647,  .647}  & \cellcolor[rgb]{ .647,  .647,  .647}\textcolor[rgb]{ 1,  0,  0}{(0.095)} & \cellcolor[rgb]{ .647,  .647,  .647} & \cellcolor[rgb]{ .647,  .647,  .647}0.056  & \cellcolor[rgb]{ .647,  .647,  .647} & \cellcolor[rgb]{ .647,  .647,  .647}0.076  & \cellcolor[rgb]{ .647,  .647,  .647} & \cellcolor[rgb]{ .647,  .647,  .647}0.084  &\cellcolor[rgb]{ .647,  .647,  .647}  & \cellcolor[rgb]{ .647,  .647,  .647}\textcolor[rgb]{ 1,  0,  0}{(0.055)} & \cellcolor[rgb]{ .647,  .647,  .647} & \cellcolor[rgb]{ .647,  .647,  .647}0.044  & \cellcolor[rgb]{ .647,  .647,  .647} & \cellcolor[rgb]{ .647,  .647,  .647}0.084  \\
    \midrule
    \midrule
    \multicolumn{2}{c}{\multirow{6}[2]{*}{SAM$^3$}} & ViT\_B & 0.564  &    & 0.016  &    & 0.726  &    & 0.564  &    & 0.566  &    & 0.017  &    & 0.729  &    & 0.566  \\
    \multicolumn{2}{c}{} & \cellcolor[rgb]{ .647,  .647,  .647}Difference($\Delta$) & \cellcolor[rgb]{ .647,  .647,  .647}0.178  & \cellcolor[rgb]{ .647,  .647,  .647}  & \cellcolor[rgb]{ .647,  .647,  .647}\textcolor[rgb]{ 1,  0,  0}{(0.206)} & \cellcolor[rgb]{ .647,  .647,  .647} & \cellcolor[rgb]{ .647,  .647,  .647}0.148  & \cellcolor[rgb]{ .647,  .647,  .647} & \cellcolor[rgb]{ .647,  .647,  .647}0.178  & \cellcolor[rgb]{ .647,  .647,  .647} & \cellcolor[rgb]{ .647,  .647,  .647}0.170  &  \cellcolor[rgb]{ .647,  .647,  .647} & \cellcolor[rgb]{ .647,  .647,  .647}\textcolor[rgb]{ 1,  0,  0}{(0.061)} & \cellcolor[rgb]{ .647,  .647,  .647}  & \cellcolor[rgb]{ .647,  .647,  .647}0.083  & \cellcolor[rgb]{ .647,  .647,  .647} & \cellcolor[rgb]{ .647,  .647,  .647}0.170  \\
    \multicolumn{2}{c}{} & ViT\_L & 0.706  &    & 0.017  &    & 0.793  &    & 0.706  &    & 0.700  &    & 0.018  &    & 0.789  &    & 0.700  \\
    \multicolumn{2}{c}{} & \cellcolor[rgb]{ .647,  .647,  .647}Difference($\Delta$) & \cellcolor[rgb]{ .647,  .647,  .647}0.236  & \cellcolor[rgb]{ .647,  .647,  .647} & \cellcolor[rgb]{ .647,  .647,  .647}\textcolor[rgb]{ 1,  0,  0}{(0.113)} & \cellcolor[rgb]{ .647,  .647,  .647} & \cellcolor[rgb]{ .647,  .647,  .647}0.130  & \cellcolor[rgb]{ .647,  .647,  .647}  & \cellcolor[rgb]{ .647,  .647,  .647}0.236  & \cellcolor[rgb]{ .647,  .647,  .647} & \cellcolor[rgb]{ .647,  .647,  .647}0.235  &  \cellcolor[rgb]{ .647,  .647,  .647} & \cellcolor[rgb]{ .647,  .647,  .647}\textcolor[rgb]{ 1,  0,  0}{(0.057)} & \cellcolor[rgb]{ .647,  .647,  .647}  & \cellcolor[rgb]{ .647,  .647,  .647}0.104  &\cellcolor[rgb]{ .647,  .647,  .647}  & \cellcolor[rgb]{ .647,  .647,  .647}0.235  \\
    \multicolumn{2}{c}{} & ViT\_H & 0.628  &    & 0.027  &    & 0.767  &    & 0.628  &    & 0.622  &    & 0.028  &    & 0.763  &    & 0.622  \\
    \multicolumn{2}{c}{} & \cellcolor[rgb]{ .647,  .647,  .647}Difference($\Delta$) & \cellcolor[rgb]{ .647,  .647,  .647}0.108  & \cellcolor[rgb]{ .647,  .647,  .647}  & \cellcolor[rgb]{ .647,  .647,  .647}\textcolor[rgb]{ 1,  0,  0}{(0.099)} &  \cellcolor[rgb]{ .647,  .647,  .647} & \cellcolor[rgb]{ .647,  .647,  .647}0.077  & \cellcolor[rgb]{ .647,  .647,  .647}  & \cellcolor[rgb]{ .647,  .647,  .647}0.108  & \cellcolor[rgb]{ .647,  .647,  .647} & \cellcolor[rgb]{ .647,  .647,  .647}0.117  &  \cellcolor[rgb]{ .647,  .647,  .647} & \cellcolor[rgb]{ .647,  .647,  .647}\textcolor[rgb]{ 1,  0,  0}{(0.059)} & \cellcolor[rgb]{ .647,  .647,  .647}  & \cellcolor[rgb]{ .647,  .647,  .647}0.065  & \cellcolor[rgb]{ .647,  .647,  .647}  & \cellcolor[rgb]{ .647,  .647,  .647}0.117  \\
\cmidrule{1-2}    \end{tabular}%
  \label{tab:cq}%
\end{table}

\subsection{Qualitative Comparison}

\begin{figure}[!htbp]
\centering
\includegraphics[width=0.7\linewidth]{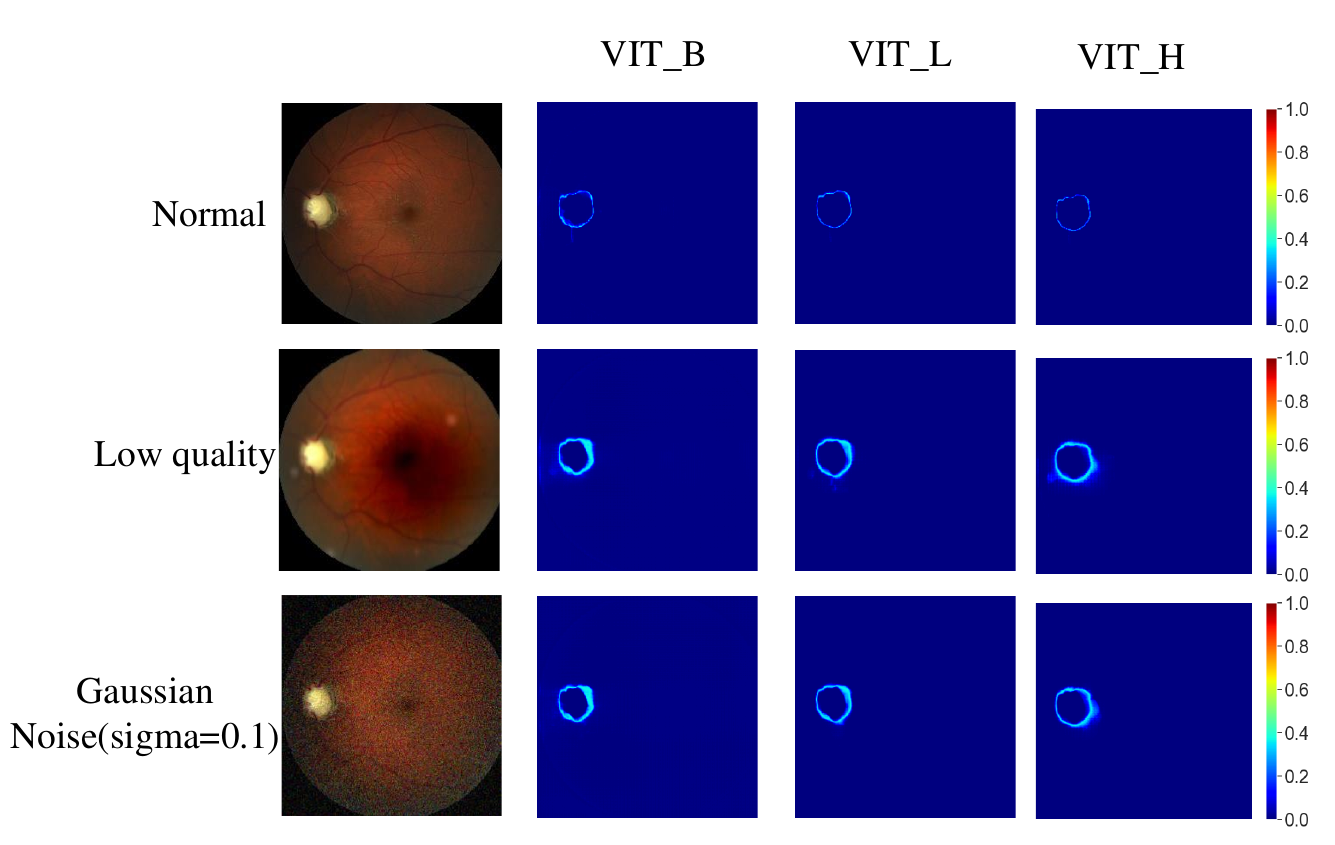}
\caption{Uncertainty distributions of the same sample under different states estimated by multi-box prompts.}
\label{F_4}
\end{figure}

\begin{figure}[!htbp]
\centering
\includegraphics[width=0.8\linewidth]{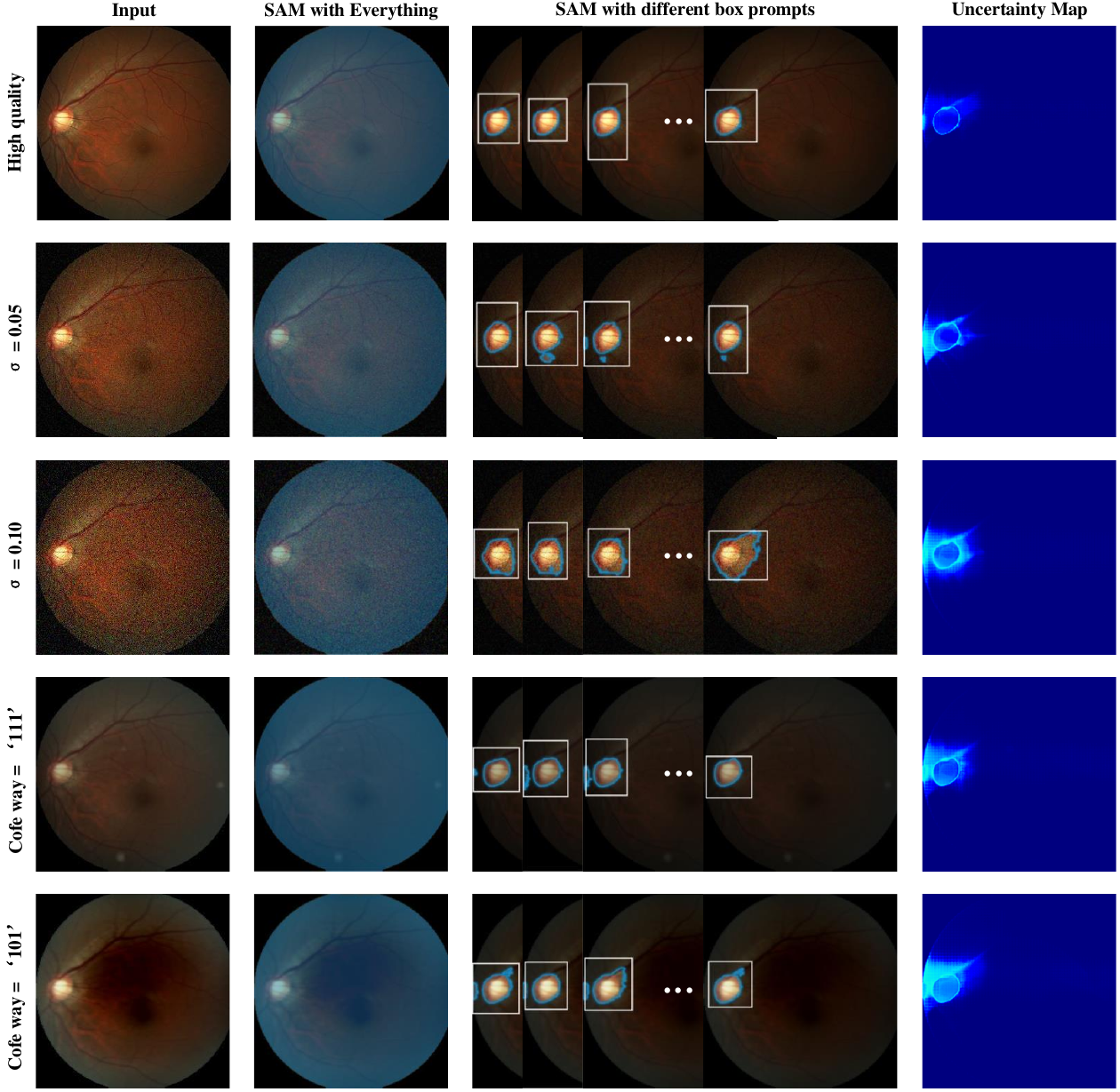}
\caption{Qualitative results of different quality fundus images in SAM's everything and box modes.}
\label{F_5}
\end{figure}

As shown in Fig.~\ref{F_4}, we first show the uncertainty estimation results under SAM with multi-box mode. As can be seen from it, the periphery of the eye disc is clearly marked as an area of uncertainty.  Furthermore, we compare the segmentation results of different modes of SAM under normal and degraded medical images, as shown in Fig.~\ref{F_5}. In SAM with everything mode, it is difficult to segment the eye disc. Under the box prompt, the eye disc can be segmented under normal conditions, but the results under Gaussian noise and degraded images are not satisfactory. While our method also achieves better segmentation results in degraded images and provides weights for uncertain pixels. This opens a new paradigm for SAM towards robust and reliable medical image segmentation.

\section{Discussion and Conclusion}
In this paper, we investigated the segmentation performance of SAM on fundus images. The results have shown that box prompt significantly improve the segmentation, but different box prompts lead to variations in predictions. The main method proposed in this paper, prompt augmentation, can help estimate the variations by aleatoric uncertainty and produce an uncertainty distribution map that highlights challenging areas for segmentation. The uncertainty map not only improves the segmentation process and final results but also enables the development of more advanced methods for segmenting fundus images. Moreover, the uncertainty map offers valuable guidance in areas where manual annotation is required. The feature of using the uncertainty distribution map for guiding segmentation and improving accuracy is noteworthy. Furthermore, the uncertainty map can help identify potential segmentation errors and support further analysis, providing useful information for clinicians.


%



%
\bibliographystyle{IEEEtran}
\bibliography{SAMU.bib}

\begin{thebibliography}{10}
\providecommand{\url}[1]{#1}
\csname url@samestyle\endcsname
\providecommand{\newblock}{\relax}
\providecommand{\bibinfo}[2]{#2}
\providecommand{\BIBentrySTDinterwordspacing}{\spaceskip=0pt\relax}
\providecommand{\BIBentryALTinterwordstretchfactor}{4}
\providecommand{\BIBentryALTinterwordspacing}{\spaceskip=\fontdimen2\font plus
\BIBentryALTinterwordstretchfactor\fontdimen3\font minus
  \fontdimen4\font\relax}
\providecommand{\BIBforeignlanguage}[2]{{%
\expandafter\ifx\csname l@#1\endcsname\relax
\typeout{** WARNING: IEEEtran.bst: No hyphenation pattern has been}%
\typeout{** loaded for the language `#1'. Using the pattern for}%
\typeout{** the default language instead.}%
\else
\language=\csname l@#1\endcsname
\fi
#2}}
\providecommand{\BIBdecl}{\relax}
\BIBdecl

\bibitem{floridi2020gpt}
L.~Floridi and M.~Chiriatti, ``Gpt-3: Its nature, scope, limits, and
  consequences,'' \emph{Minds and Machines}, vol.~30, pp. 681--694, 2020.

\bibitem{kirillov2023segment}
A.~Kirillov, E.~Mintun, N.~Ravi, H.~Mao, C.~Rolland, L.~Gustafson, T.~Xiao,
  S.~Whitehead, A.~C. Berg, W.-Y. Lo \emph{et~al.}, ``Segment anything,''
  \emph{arXiv preprint arXiv:2304.02643}, 2023.

\bibitem{tang2023can}
L.~Tang, H.~Xiao, and B.~Li, ``Can sam segment anything? when sam meets
  camouflaged object detection,'' \emph{arXiv preprint arXiv:2304.04709}, 2023.

\bibitem{ji2023sam}
G.-P. Ji, D.-P. Fan, P.~Xu, M.-M. Cheng, B.~Zhou, and L.~Van~Gool, ``Sam
  struggles in concealed scenes--empirical study on" segment anything",''
  \emph{arXiv preprint arXiv:2304.06022}, 2023.

\bibitem{ji2023segment}
W.~Ji, J.~Li, Q.~Bi, W.~Li, and L.~Cheng, ``Segment anything is not always
  perfect: An investigation of sam on different real-world applications,''
  \emph{arXiv preprint arXiv:2304.05750}, 2023.

\bibitem{he2023can}
H.~Sheng, B.~Rina, L.~Jingpeng, G.~P.~Ellen, and O.~Yangming, ``Accuracy of
  segment-anything model (sam) in medical image segmentation tasks,''
  \emph{arXiv preprint arXiv:2304.09324}, 2023.

\bibitem{roy2023sam}
S.~Roy, T.~Wald, G.~Koehler, M.~R. Rokuss, N.~Disch, J.~Holzschuh, D.~Zimmerer,
  and K.~H. Maier-Hein, ``Sam. md: Zero-shot medical image segmentation
  capabilities of the segment anything model,'' \emph{arXiv preprint
  arXiv:2304.05396}, 2023.

\bibitem{16dropout}
Y.~Gal and Z.~Ghahramani, ``Dropout as a bayesian approximation: Representing
  model uncertainty in deep learning,'' in \emph{international conference on
  machine learning}.\hskip 1em plus 0.5em minus 0.4em\relax PMLR, 2016, pp.
  1050--1059.

\bibitem{zou2022tbrats}
K.~Zou, X.~Yuan, X.~Shen, M.~Wang, and H.~Fu, ``Tbrats: Trusted brain tumor
  segmentation,'' in \emph{International Conference on Medical Image Computing
  and Computer-Assisted Intervention}.\hskip 1em plus 0.5em minus 0.4em\relax
  Springer, 2022, pp. 503--513.

\bibitem{zou2023evidencecap}
K.~Zou, X.~Yuan, X.~Shen, Y.~Chen, M.~Wang, R.~S.~M. Goh, Y.~Liu, and H.~Fu,
  ``Evidencecap: Towards trustworthy medical image segmentation via evidential
  identity cap,'' \emph{arXiv preprint arXiv:2301.00349}, 2023.

\bibitem{li2022region}
H.~Li, Y.~Nan, J.~Del~Ser, and G.~Yang, ``Region-based evidential deep learning
  to quantify uncertainty and improve robustness of brain tumor segmentation,''
  \emph{Neural Computing and Applications}, pp. 1--15, 2022.

\bibitem{ICMLdeterministic20}
J.~Van~Amersfoort, L.~Smith, Y.~W. Teh, and Y.~Gal, ``Uncertainty estimation
  using a single deep deterministic neural network,'' in \emph{International
  Conference on Machine Learning}.\hskip 1em plus 0.5em minus 0.4em\relax PMLR,
  2020, pp. 9690--9700.

\bibitem{2020NIPSdeterministic}
J.~Z. Liu, Z.~Lin, S.~Padhy, D.~Tran, T.~Bedrax-Weiss, and B.~Lakshminarayanan,
  ``Simple and principled uncertainty estimation with deterministic deep
  learning via distance awareness,'' in \emph{Proceedings of the 34th
  International Conference on Neural Information Processing Systems}, 2020.

\bibitem{roy2019bayesian}
A.~G. Roy, S.~Conjeti, N.~Navab, C.~Wachinger, A.~D.~N. Initiative
  \emph{et~al.}, ``Bayesian quicknat: Model uncertainty in deep whole-brain
  segmentation for structure-wise quality control,'' \emph{NeuroImage}, vol.
  195, pp. 11--22, 2019.

\bibitem{ensemble17}
B.~Lakshminarayanan, A.~Pritzel, and C.~Blundell, ``Simple and scalable
  predictive uncertainty estimation using deep ensembles,'' \emph{Advances in
  Neural Information Processing Systems}, vol.~30, 2017.

\bibitem{17dropoutCV}
A.~Kendall and Y.~Gal, ``What uncertainties do we need in bayesian deep
  learning for computer vision?'' in \emph{NIPS}, 2017.

\bibitem{MIA2020exploringDropSeg}
T.~Nair, D.~Precup, D.~L. Arnold, and T.~Arbel, ``Exploring uncertainty
  measures in deep networks for multiple sclerosis lesion detection and
  segmentation,'' \emph{Medical image analysis}, vol.~59, p. 101557, 2020.

\bibitem{wang2018TTA}
G.~Wang, W.~Li, S.~Ourselin, and T.~Vercauteren, ``Automatic brain tumor
  segmentation using convolutional neural networks with test-time
  augmentation,'' in \emph{International MICCAI Brainlesion Workshop}.\hskip
  1em plus 0.5em minus 0.4em\relax Springer, 2018, pp. 61--72.

\bibitem{fu2018joint}
H.~Fu, J.~Cheng, Y.~Xu, D.~W.~K. Wong, J.~Liu, and X.~Cao, ``Joint optic disc
  and cup segmentation based on multi-label deep network and polar
  transformation,'' \emph{IEEE transactions on medical imaging}, vol.~37,
  no.~7, pp. 1597--1605, 2018.

\bibitem{bein2006entropy}
B.~Bein, ``Entropy,'' \emph{Best Practice \& Research Clinical
  Anaesthesiology}, vol.~20, no.~1, pp. 101--109, 2006.

\bibitem{orlando2020refuge}
J.~I. Orlando, H.~Fu, J.~B. Breda, K.~Van~Keer, D.~R. Bathula, A.~Diaz-Pinto,
  R.~Fang, P.-A. Heng, J.~Kim, J.~Lee \emph{et~al.}, ``Refuge challenge: A
  unified framework for evaluating automated methods for glaucoma assessment
  from fundus photographs,'' \emph{Medical image analysis}, vol.~59, p. 101570,
  2020.

\bibitem{deep_reitna_enhance}
Z.~Shen, H.~Fu, J.~Shen, and L.~Shao, ``Modeling and enhancing low-quality
  retinal fundus images,'' \emph{IEEE Transactions on Medical Imaging},
  vol.~40, no.~3, pp. 996--1006, 2020.

\bibitem{guo2017calibration}
C.~Guo, G.~Pleiss, Y.~Sun, and K.~Q. Weinberger, ``On calibration of modern
  neural networks,'' in \emph{International conference on machine
  learning}.\hskip 1em plus 0.5em minus 0.4em\relax PMLR, 2017, pp. 1321--1330.

\bibitem{fan2017structure}
D.-P. Fan, M.-M. Cheng, Y.~Liu, T.~Li, and A.~Borji, ``Structure-measure: A new
  way to evaluate foreground maps,'' in \emph{Proceedings of the IEEE
  international conference on computer vision}, 2017, pp. 4548--4557.

\bibitem{margolin2014evaluate}
R.~Margolin, L.~Zelnik-Manor, and A.~Tal, ``How to evaluate foreground maps?''
  in \emph{Proceedings of the IEEE conference on computer vision and pattern
  recognition}, 2014, pp. 248--255.

\end{thebibliography}

\end{document}